\documentclass[letterpaper, 10 pt, conference]{ieeeconf}  

\IEEEoverridecommandlockouts                              

\usepackage{amsmath}
\usepackage{amsfonts}
\usepackage{graphicx}

\pdfoutput=1

\title{\LARGE \bf
A Graph Based Neural Network Approach to Immune Profiling of Multiplexed Tissue Samples
}

\author{Natalia Garcia Martin $^{1, 2}$, Stefano Malacrino $^{1, 3}$, Marta Wojciechowska $^{1, 4}$, Leticia Campo $^{4}$, \\ Helen Jones $^{5}$, David C. Wedge $^{6}$, Chris Holmes $^{1, 2}$, Korsuk Sirinukunwattana $^{1, 4}$, Heba Sailem $^{1, 4}$, \\Clare Verrill $^{3, 7}$, and Jens Rittscher $^{1, 4, 7}$
\thanks{$^{1}$Big Data Institute, University of Oxford, Li Ka Shing Centre for Health
Information and Discovery, Oxford, UK
        {\tt\small natalia.garciamartin@spc.ox.ac.uk}}%
\thanks{$^{2}$Department of Statistics, University of Oxford, Oxford, UK}%
\thanks{$^{3}$Nuffield Department of Surgical Sciences, University of Oxford, Oxford, UK}%
\thanks{$^{4}$Institute of Biomedical Engineering, Department of Engineering Science,
University of Oxford, Oxford, UK,}%
\thanks{$^{5}$Department of Colorectal Surgery, Oxford University Hospitals NHS Trust, Oxford, UK}%
\thanks{$^{6}$Manchester Cancer Research Centre, University of Manchester, Manchester, UK}%
\thanks{$^{7}$NIHR Oxford Biomedical Research Centre, Oxford, UK}%
\thanks{*NGM is supported by Cancer Research UK (CRUK), through a CRUK Oxford Centre Prize DPhil Studentship (C2195/A27450). MW is funded by the UK Engineering and Physical Sciences Research Council and Medical Research Council (EP/L016052/1) and in part by Perspectum Ltd. SM and JR are supported by the PathLAKE Centre of Excellence for digital pathology and artificial intelligence which is funded from the Data to Early Diagnosis and Precision Medicine strand of the HM Government’s Industrial Strategy Challenge Fund, managed and delivered by Innovate UK on behalf of UK Research and Innovation (UKRI). Views expressed are those of the authors and not necessarily those of the PathLAKE Consortium members, the NHS, Innovate UK or UKRI. Grant ref: File Ref 104689/ application number 18181. JR is in part funded by the National Institute for Health Research Oxford Biomedical Research Centre.}}

\begin{document}

\maketitle
\thispagestyle{empty}
\pagestyle{empty}

\begin{abstract}

Multiplexed immunofluorescence provides an unprecedented opportunity for studying specific cell-to-cell and cell microenvironment interactions. We employ graph neural networks to combine features obtained from tissue morphology with measurements of protein expression to  profile the tumour microenvironment associated with different tumour stages. Our framework presents a new approach to analysing and processing these complex multi-dimensional datasets that overcomes some of the key challenges in analysing these data and opens up the opportunity to abstract biologically meaningful interactions.
\newline
\end{abstract}

\section{Introduction}
\label{sec:intro}

Novel tissue multiplexing imaging platforms \cite{stack2014multiplexed,bodenmiller2016multiplexed} allow the analysis of a broad range of cell types in the tissue architecture context. These approaches open up new opportunities for improving our understanding of disease, monitoring therapeutic response, and the development of high-dimensional clinical tests. Here, we are interested in profiling the complex interaction between the tumour and the immune system within the tumour microenvironment (TME), which dictates the tumour progression. While current cancer classification highly relies on the extent of the primary tumour (T), lymph node involvement (N) and metastatic presence (M), visualising multiple protein targets in the same tissue allows us to interrogate the role of adaptive immune cell infiltration in colorectal cancer (CRC) prognosis.

The analysis of multiplexing data requires the combination of spatial information that captures the changes in tissue architecture with measurements of protein expression. When compared to standard digital pathology, multiplexing datasets are typically much smaller and contain imaging artifacts and strong variations in protein expression, making this a particularly challenging problem. Moreover, interpretability is a key aspect when working with multiplexed data to be able to link the analysis to any underlying biological hypothesis.

Building on recent success of applying graph neural networks (GNNs) to histopathology, we introduce a novel framework for analysing multiplexed immunofluorescence (IF) images using GNNs. Constructing graphs from multiplexed IF data is non-trivial due to the stated challenges. Our approach overcomes these challenges by: (1) including a selection of network metrics that capture the interactions between the immune cells and the tumour; (2) a hierarchical structure that considers both the cell-level and the spatial tissue arrangement; (3) implicit denoising from the use of message-passing; (4) data augmentation on the graphs to account for the limited amount of training data; and (5) the opportunity to interpret results in order to identify the tissue areas contributing the most to the predictions. 
In summary, we propose a GNN model to  profile the tumour microenvironment associated with different tumour stages in an explainable setting.

\section{Methods}
\label{sec:methods}

\begin{figure*}[t]
\centering
\includegraphics[width=14cm]{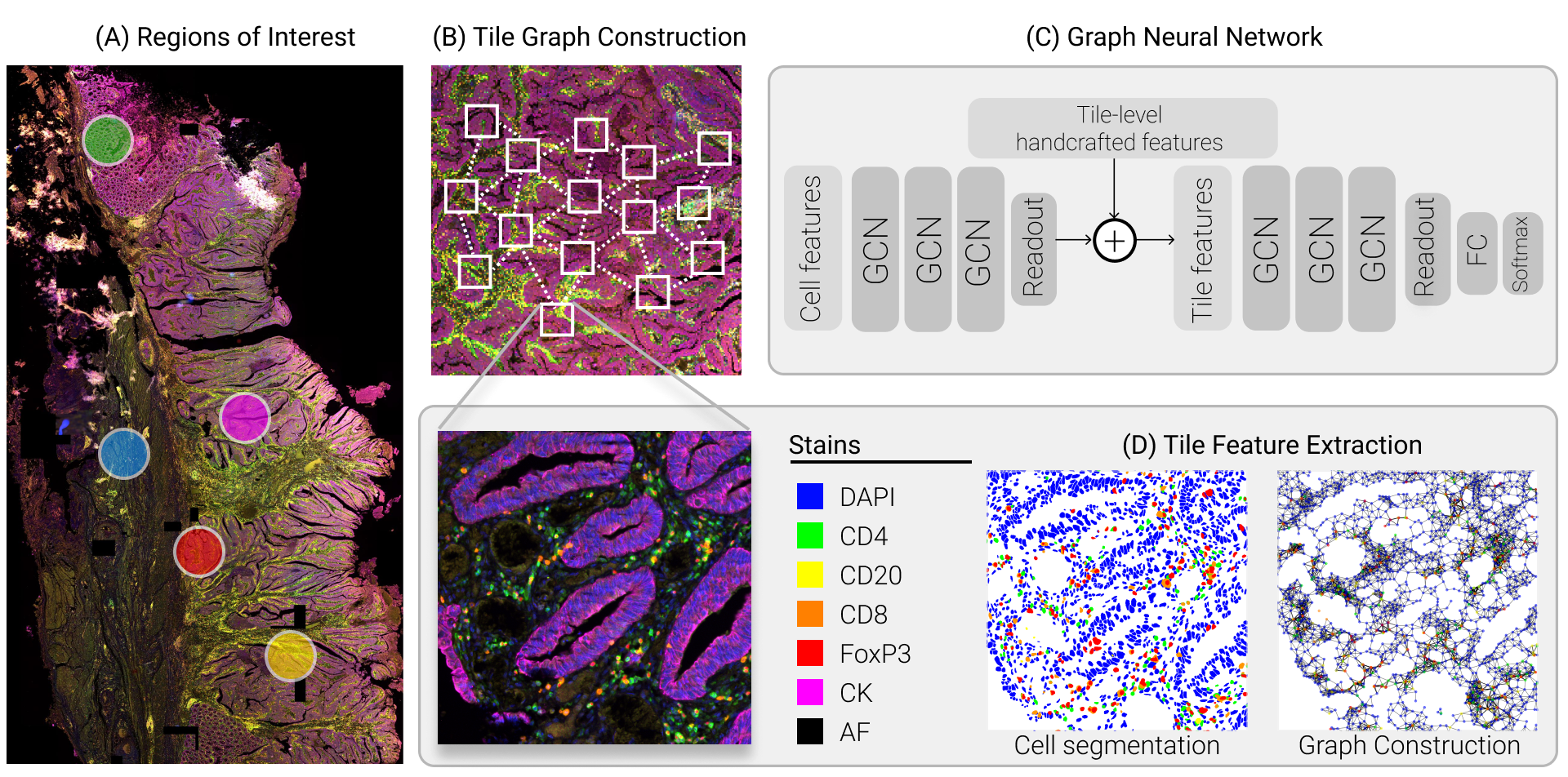}
\caption{Overview of the proposed method described in Section \ref{sec:methods}. The regions of interest correspond to the tumour centre (yellow and magenta), invasive tumour front (red), background mucosa (green), and peritumoural stroma (blue).} 
\label{f:overview}
\end{figure*}

Figure \ref{f:overview} provides a summary of the overall approach. Prior to the analysis, we carefully pre-process the data by applying fluorescent image correction algorithms and subsequently identify cell nuclei using a robust segmentation approach. Rather than performing a global analysis of the slide, we perform a local analysis in selected regions of interest (RoIs)(see Fig. \ref{f:overview}A). The two-layer graph described in Section \ref{ss:graph-construction} is constructed to abstract the key biological interactions of the underlying tissue. It first captures the location of cells, certain morphological measurements and protein expressions to form a cell-graph (Fig. \ref{f:overview}D). After message-passing, the updated cell embeddings are aggregated at the tile level and concatenated with the set of hand-crafted immune-interaction features that we describe in Section \ref{cell-graph-feats}. The second set of graphs are constructed at the RoI level, with nodes representing tile centroids to form a tile-graph (Fig. \ref{f:overview}B). These RoI-level graphs are then fed into the model for pT stage prediction (Fig. \ref{f:overview}C). Finally,
post-hoc explainability methods, presented in Section \ref{ss:post-hoc-explainability}, are utilised to visualise the relationship between immune interaction profiles and prediction of tumour stage. 

\subsection{Multiplexed IF data}

The Perkin-Elmer Vectra platform features an immune panel consisting of six fluorescent markers. DAPI is used for nuclei segmentation. Cytokeratin is used to delineate epithelial cells. A further four markers are included to depict immune cells: CD4 for helper T-cells, CD8 for cytotoxic T-cells, CD20 for B-cells, and Foxp3 for regulatory T-cells. The system also provides a seventh channel corresponding to the imaging system's autofluorescence isolation capacity which improves signal-to-noise ratio \cite{inform}. 

\subsection{Nuclei segmentation and cell phenotyping} 
Multiplexed image data requires careful preprocessing as tiles are scanned independently. Hence before segmentation, a background and shading correction \cite{peng2017basic} was performed to improve the stitching of individual image tiles. Moreover, contrast limited adaptive histogram equalisation (CLAHE) was used to improve contrast in the DAPI channel, which is used for segmentation. The segmentation network employed to identify cell nuclei consists of a 3-class (cell inside, cell boundary, background) modified U-Net architecture comprised of the original U-Net \cite{ronneberger2015u} decoder and a ResNet \cite{he2016deep} encoder. The model was pre-trained on the fluorescence samples from the publicly available BBBC038v1 dataset \cite{ljosa2012annotated} to ensure the correct identification of cells of different sizes and to ignore DAPI positive fragments not corresponding to actual cells. Segmentation masks were projected onto the remaining channels to measure the average nuclei protein expression. Cells were then assigned to the cell type corresponding to the marker whose expression was located in the highest percentile rank.

\subsection{Two-layer graph representation}
\label{ss:graph-representation}

After image segmentation we define an undirected graph $G=(V, E)$, with vertices $V$ and edges $E$. 
Similar to Pati {\it et al.} \cite{pati2020hact} we employ a two-layer graph representation, with (1) cell-graphs \cite{gunduz2004cell} constructed on small randomly sampled tiles, where nodes represent nuclei centroids in order to quantify local patterns of immune interaction, and (2) a tile-level graph able to aggregate information from the multiple tissue regions. The graph topology is represented by an adjacency matrix $A \in {\mathbb{R}}^{N\times N}$, and node features are represented by the matrix $X \in {\mathbb{R}}^{N\times D}$, with $N=|V|$ and feature dimension $D$. We construct $A$ based on a distance threshold as follows:

\begin{equation}
\label{eq1}
A_{ij}=\left\{\begin{array}{cc}
1 & \text{if } d(i, j)<k \\
0 & \text{otherwise},
\end{array}\right.
\end{equation}
where $k$ determines which nodes in the graph are connected. The choice of $k$ is described in Section \ref{ss:graph-construction}.

\subsection{Cell-graph feature extraction}
\label{cell-graph-feats}

We calculate a total of 68 handcrafted network metrics at the cell-graph level to acquire information about the distribution of each cell population of interest. 

These include the average clustering and square clustering coefficients, the assortativity, radius, density, transitivity, and the closeness of each cell type population, as defined by Schult {\it et al.} \cite{schult2008exploring}. The ratios between each pair of immune cell densities (e.g. CD4-CD8 ratio), a known prognostic factor for cancer progression  \cite{hadrup2013effector}
are also computed. To measure the degree of mixing between tumour and immune cells, we additionally compute the ratio of immune-tumour to immune-immune interactions \cite{keren2018structured}. 

\subsection{Cell-graph and tile-graph neural network}

We employ Graph Neural Networks (GNNs) to obtain a graph representation $H \in {\mathbb{R}}^{N\times P}$ from our initial cell embeddings $H^{0}=X \in {\mathbb{R}}^{N\times D}$, where $P$ is the number of output features. Using the notation from \cite{kipf2016semi} and \cite{xu2018powerful}, we first perform a number of message passing steps to obtain the node embeddings $h_v$ for each cell $v$ in the cell-graph $CG$, which we then combine into a global cell-graph embedding $h_{CG}$ for each tile. The message passing consists of an aggregation and combination of the neighbouring nodes features. For the $k^{th}$ GNN layer:
\begin{equation}
a_{v}^{(k)}=\text {AGG}^{(k)}\left(\left\{h_{u}^{(k-1)}: u \in \mathcal{N}(v)\right\}\right)
\end{equation}
\begin{equation}
h_{v}^{(k)}=\operatorname{COMBINE}^{(k)}\left(h_{v}^{(k-1)}, a_{v}^{(k)}\right)
\end{equation}
\begin{equation}
h_{CG}=\operatorname{READOUT}\left(\left\{h_{v}^{(k)} \mid v \in CG\right\}\right),
\end{equation}
where $\mathcal{N}(v)$ denotes the set of neighbours of $v$. We use the graph convolutional network (GCN) operator defined in \cite{morris2019weisfeiler}:
\begin{equation}
    h_v^{(k)} = W^{(k)}_1 h_v^{(k-1)} + W^{(k)}_2 \sum_{u \in \mathcal{N}(v)} h_u^{(k-1)}.
\end{equation}

The updated cell-graph embeddings $h_{CG}$ are then combined with the selected network metrics $m_{CG}$ listed in the previous sub-section to define the tile-graph embeddings for each tile $t$:
\begin{equation}
h_{\mathrm{t}}^{(0)}=\operatorname{CONCAT}\left(m_{\mathrm{CG}}, h_{CG}\right).
\end{equation}
These tile embeddings $h_t$ for each tile $t$ in the tile-graph $TG$ are then updated by applying Eqs. 2-3 again, where nodes now correspond to tiles. The readout layer then combines the information from the multiple tiles to obtain the final embedding $h_{TG}$ for the RoI:
\begin{equation}
h_{TG}=\operatorname{READOUT}\left(\left\{h_{t}^{(k)} \mid t \in TG\right\}\right).
\end{equation}

\subsection{Post-hoc explainability}
\label{ss:post-hoc-explainability}

We employ Integrated Gradients (IG) \cite{sundararajan2017axiomatic} in the tile-graph to understand the significance of each tile node in predicting tumour stage. We do so by computing the IG attribution of each edge and aggregating the attributions of the edges connecting each node. The IG edge attribution is computed by comparing each edge mask with a baseline of edge weights set to zero. Since we use unweighted graphs, the initial edge weights are all one. The IG for each edge $e_i$ is computed as follows:
\begin{equation}
    \text{IG}_{e_i} = \int_{\alpha =0}^1 \frac{\partial F(x_{\alpha})}{\partial w_{e_i}} d\alpha,
\end{equation}
where $x_{\alpha}$ corresponds to the original input graph but with all edge weights set to $\alpha$, $w_{e_i}$ denotes the current edge weight, and $F(x)$ is the output of the model for an input $x$. The integral is approximated using a Gauss-Legendre quadrature. 

In order to identify the key features impacting the prediction, we further run the GNN Explainer model \cite{ying2019gnn}, which maximises the mutual information MI between the prediction of the trained model and that of the simplified
explainer model given by a subgraph $G_S$ and a subset of features $T$: $\max_{G_S, T} \text{MI}(Y, (G_S, T))$.

\begin{table*}[t]
\caption{Mean and standard deviation of RoI-level class-weighted F1-scores measured on the test set and averaged over three distinct train-test splits.} 

\centering
\begin{tabular}{|l|l|l|l|l|l|l|}
\hline
Region&GCN Mean pool&GCN Add pool&GCN Max pool &MIL Attention pool & MIL Mean pool &MLP\\
\hline
All& $58.4_{\pm1.7}$ & $61.6_{\pm4.2}$ & $61.6_{\pm3.2}$& $55.1_{\pm1.4}$&$49.5_{\pm1.6}$&$49.0_{\pm0.7}$\\
Centre&  $53.8_{\pm6.4}$ & $60.7_{\pm8.9}$ &  $58.0_{\pm0.1}$& $50.0_{\pm2.1}$&$49.7_{\pm3.4}$&$49.2_{\pm0.8}$\\
Front& $63.6_{\pm8.5}$ &$60.4_{\pm9.9}$ & {\bfseries 72.9} $_{\pm7.8}$&$54.4_{\pm5.8}$&$40.0_{\pm7.1}$&$48.9_{\pm0.7}$\\
Mucosa&  $47.5_{\pm12.6}$ & $50.2_{\pm8.1}$& $60.8_{\pm9.9}$ & $63.5_{\pm11.3}$&$46.1_{\pm9.1}$&$48.4_{\pm1.9}$\\
Stroma&  $56.6_{\pm7.4}$ & $58.9_{\pm8.5}$& $61.6_{\pm6.5}$& $57.7_{\pm3.9}$&$53.0_{\pm4.5}$&$48.4_{\pm1.2}$\\
\hline
\end{tabular}
\label{tab1}
\end{table*}

\section{Experiments}

\subsection{Dataset} 
Paraffin-embedded tissue samples of 41 rectal primary tumours were used to investigate the risk of disease progression and recurrence. 
Specialist GI pathologists reported tumour stage on matching H\&E slides: 25 of these samples were assigned a pT1 tumour stage, while 16 samples were considered to be more advanced (13 pT2, 3 pT3). 
Specific regions of interest such as those shown in Fig. \ref{f:overview}A were provided by a pathologist for the tumour centre, invasive tumour front, background mucosa, and peritumoural stroma, guided by the matched H\&E image. Annotation areas correspond to the standard 1mm diameter disk size used for biopsies and tissue microarrays (TMAs), allowing for a future integrative analysis with TMA cohorts. 

\subsection{Graph construction}
\label{ss:graph-construction}

RoIs of the size of 2048x2048 pixels corresponding to the bounding box of the disk annotations are selected to investigate immune-cell interactions across samples and regions. From each RoI, 200 256x256 tiles are randomly chosen to construct cell-graphs using NetworkX \cite{schult2008exploring}, with nodes positioned at the centroid of each nucleus. We set $k$ from Eq. \ref{eq1} to be 30 pixels. This results in a small node degree as well as a small number of disconnected nodes in order to reduce graph complexity and facilitate metric computation. 

For each node, we record the average expression for the five markers of interest (CD4, CD8, CD20, FoxP3, CK), the area occupied by the cell, and the cell solidity. These 7 features are inserted as node features. We subsequently perform three message-passing steps to update the node features by encompassing information from nearby cells, which are aggregated using mean pooling and transformed into a vector
of length 16. Additionally, for each tile, we compute the set of 68 hand-crafted immune-interaction features enumerated in Section \ref{cell-graph-feats}. Nuclei and cell-interaction features are then concatenated into a vector of length 84 per tile. The second set of graphs are constructed at the RoI level, with nodes representing the 200 sampled tiles positioned at their tile centroids and node features corresponding to the selected 84 attributes. These RoI-level graphs are then fed into the model for pT stage prediction.

\subsection{Data augmentation}

The augmented set is obtained by constructing the networks using a subset of 80\% of the nodes at each step (160 tiles) and by varying the threshold $k$ that needs to be surpassed for an edge to be included between two neighbours by sampling a value in the pixel range \{150, 175, 200, 225, 250\}, resulting in a variety of tighter and sparser graphs. The node subsampling and edge modifications ensure that networks in the training set are sufficiently different to avoid over-fitting. For the test set, only a single network is constructed per RoI using the default distance threshold of 200 pixels for adjacency construction and the full set of tile nodes in the RoI (200 tiles).

\subsection{Implementation}
The model consists of three GraphConv \cite{morris2019weisfeiler,FeyLenssen2019} layers with ReLu activation and global pooling aggregation. Experiments are conducted in PyTorch 1.7.0 using PyTorch Geometric \cite{FeyLenssen2019}.

Data are split into training and testing at the patient level. We use 70\% (134) of the RoIs for training and 30\% (59) for testing. Due to the limited sample size, a pseudo-validation set is constructed by randomly sampling (with pT stratification) 10\% of the pre-augmented training data, and used for hyperparameter grid-search. Model performance is measured according to their weighted F1-scores on the test set. The model is trained using an L2-regularised Adam optimiser and a weighted cross entropy loss. We employed early stopping based on the pseudo-validation set.
We tune the hyperparameters using a grid search. The values that provide the best performance in terms of class-weighted F1 score correspond to a dropout ratio of $0.5$, a learning rate of 10e-5, a weight decay of 10e-5, $32$ hidden layers, and a batch size of $64$. 

Two baseline models are considered: the first one consists of a multi-layer perceptron (MLP) which takes as input the average individual nuclei features (size, shape and marker expression) without taking the cell topology into consideration. The second consists of a multi-instance learning (MIL) approach which takes as input both the cell-level features and the cell-graph features, computed as in section \ref{cell-graph-feats}. We consider this model with both attention and average pooling.

\subsection{Post-hoc explainability}

We compute Integrated Gradients \cite{sundararajan2017axiomatic} using the model interpretability library for PyTorch Captum \cite{kokhlikyan2020captum} to obtain an importance score of individual edges and nodes for the pT stage prediction of each instance in the test set. We can then compare areas of predictive importance across the different selected RoI regions. The GNN Explainer model (implemented using PyTorch Geometric) is used to obtain feature importances across all tiles in the test set.

\begin{figure}[t]
\centering
\includegraphics[width=8.5cm]{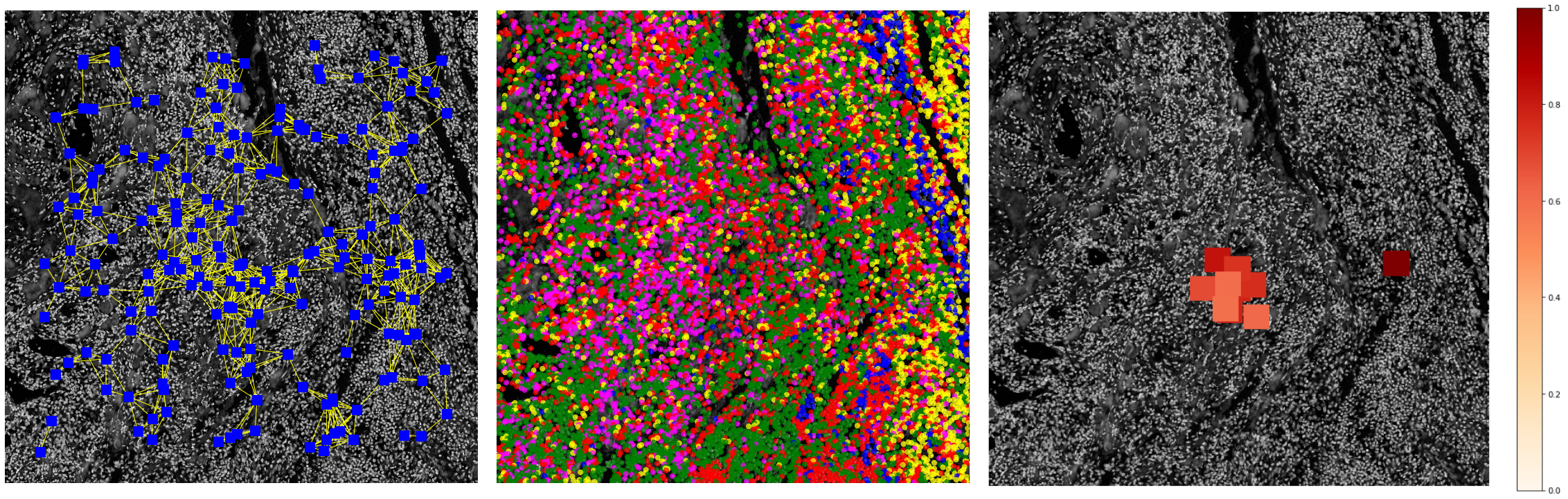}
\caption{An example of an invasive front RoI for a pT0 sample classified correctly. (Left) Tile-graph of 200 256x256 tiles overlaid on DAPI. (Centre) Cell-graph corresponding to the 2048x2048 RoI: blue - epithelial, green - T-helper, red: cytotoxic T-cell, magenta: T-reg, yellow: B-cell. (Right) Top ten tiles classified as important using integrated gradients for predicting tumour stage from immune interaction features.}
\label{fig:explain}
\end{figure}

\subsection{Results and discussion}
As shown in Table \ref{tab1}, in the majority of the graph-based experiments the invasive front was the region with the highest predictive power, followed by the peritumoural stroma, known to have a high prognostic impact. Moreover, all the graph-based models present an improvement over the baseline models: this result suggests that the network topology plays an important role in tumour stage classification.
Among the graph-based models, global max-pooling performed better than average pooling. Due to the limited number of samples with pT3, the classification of these RoIs was challenging. However, the majority of these RoIs were predicted to have pT2 stage, demonstrating that the model has learned to identify immune features related to an advanced cancer state. The proportion of interactions between CD4+ and CD8+ cells, the interactions between FoxP3 positive and epithelial cells, and the average expression of CD20 were determined by GNN Explainer as the top three features affecting tumour stage classification. Fig. \ref{fig:explain} shows an example of tiles selected by IG as important in an invasive front RoI: it can be observed that the network considers a large cluster of regulatory T-cells as the most significant area for the prediction. 

\section{CONCLUSIONS}

Our experiments demonstrate that the proposed two-layer GNN opens up new possibilities for interrogating multiplexed immuno-fluorescence images. As the model is capable of predicting tumour stage with an F1-score above 60\%, we conclude that the model captures disease relevant information at a local level. The improvement over the baseline observed with the models that use GCNs, which are able to capture more complex spatial interactions, suggests that effectively modeling the cell topology plays an important role in the tumour stage classification. However, the improvement in performance is not the only advantage gained through the use of the proposed method. First, we were able to naturally denoise the marker expressions of the cells by means of the message-passing steps. Second, our hierarchical graph structure generated biologically meaningful entities which would not have been otherwise acquired through the use of convolutional neural networks. Third, by applying post-hoc explainability methods on the tile-graph we were able to identify the regions that contributed the most to the classification hence profiling local interaction patterns. This will enable a follow up analysis to identify explicit entities at the cellular and tissue level that are of biological and clinical interest.

\addtolength{\textheight}{-12cm}   




\section*{ACKNOWLEDGMENT}

The authors thank Avelino Javer for the development and training of the cell segmentation algorithm.


\end{document}